\documentclass[conference]{IEEEtran}
\IEEEoverridecommandlockouts
\usepackage{cite}
\usepackage{amsmath,amssymb,amsfonts}

\usepackage{graphicx}
\usepackage{textcomp}
\usepackage{xcolor}
\usepackage{amsmath}
\usepackage{dsfont}
\usepackage{booktabs}
\usepackage{tcolorbox}
\usepackage{algorithm}
\usepackage{multicol}
\usepackage{bbm}
\usepackage{multirow}
\usepackage{algpseudocode}  
\usepackage{CJKutf8}
\usepackage{pinyin}
\usepackage{url}
\tcbuselibrary{breakable}

\newcommand{\chinese}[1]{%
  \begin{CJK}{UTF8}{gbsn}#1\end{CJK}%
}
\newcommand{\pinyin}[1]{\textit{#1}}
\def\BibTeX{{\rm B\kern-.05em{\sc i\kern-.025em b}\kern-.08em
    T\kern-.1667em\lower.7ex\hbox{E}\kern-.125emX}}
\begin{document}

\title{PERL: Pinyin Enhanced Rephrasing Language Model for Chinese ASR N-best Error Correction
}

\author{
\IEEEauthorblockN{Junhong Liang}
\IEEEauthorblockA{
\textit{Mohamed bin Zayed University of Artificial Intelligence} \\
Abu Dhabi, United Arab Emirates \\
junhong.liang@mbzuai.ac.ae
}
\and
\IEEEauthorblockN{Bojun Zhang}
\IEEEauthorblockA{
\textit{Institute of Automation, Chinese Academy of Sciences} \\
Beijing, China \\
zhangbojun2022@ia.ac.cn
}
}

\maketitle

\begin{abstract}
Chinese ASR correction is challenging because errors are often \emph{phonetic} (many characters share similar Pinyin) while the correction model must also obey a \emph{length constraint} under noisy N-best hypotheses. Existing approaches either exploit Pinyin only at the prompt/feature level without integrating it into model representations or rely on generative decoding that can drift in length.
We propose \textbf{PERL}, a \textbf{constrained rephrasing pipeline} for Chinese N-best ASR correction that (i) predicts the target length and enforces it via mask budgeting, and (ii) fuses \emph{semantic} and \emph{phonetic} (Pinyin) representations through token-wise gates conditioned on sentence semantics. Experiments on Aishell-1 and our new domain N-best benchmark \textbf{DoAD} show that PERL consistently reduces CER (29.11\% on Aishell-1 and up to $\sim$70\% on DoAD) while maintaining low latency. We also provide analyzes of length generalization and phonetic--semantic interactions, showing when PERL relies on phonetic cues versus semantic constraints.
\end{abstract}


\section{Introduction}
Automatic Speech Recognition (ASR) has been widely adopted in applications such as voice interaction and information retrieval. Despite advances in end-to-end models \cite{radford2022robustspeechrecognitionlargescale}, ASR output remains vulnerable to accents, background noise, and speaker variability. These errors not only reduce recognition quality, but also propagate to downstream tasks, motivating the need for practical post-correction modules.

In Mandarin Chinese, ASR errors often follow phonetic patterns, as many characters share similar pronunciations. This is closely related to the Chinese phonetic system, \emph{Pinyin}, a Romanized representation of character pronunciations. Previous work on Chinese Spelling Correction (CSC) has demonstrated that incorporating phonological features improves error correction \cite{xu_read_2021,li-etal-2022-improving-chinese,wu-wu-2022-spelling, liang-etal-2024-hybrid}. Extending this idea to ASR, recent studies have integrated Pinyin features as explicit prompts for LLMs \cite{tang_pinyin_2024}. However, to bridge the gap between text and phonetic modalities, it is essential to incorporate phonological cues directly into model representations rather than relying solely on explicit prompts. 

Existing approaches are largely dependent on limited embeddings or generative models, which cannot guaranty fixed-length outputs and are thus less suitable for ASR correction \cite{xing2024mitigating,li2024cllmlearncheckchinese,liang2025rairretrievalaugmentediterativerefinement}. Although large language models (LLMs) have achieved strong results in CSC \cite{li2023ineffectivenesslargelanguagemodels}, their decoder-only structure tends to produce predictions of variable-length. Unlike grammatical error correction, where flexible generation is acceptable \cite{zhang_mucgec_2022}, or spelling correction, which typically assumes fixed-length outputs \cite{zhang2020spelling}, ASR correction requires a constrained setting in which the model must infer the correct sentence length from noisy hypotheses. Existing methods have explored constrained decoding \cite{Ma_2023_NbestT5}, alignment algorithms \cite{leng-etal-2021-fastcorrect-2}, and LLM-based re-ranking \cite{pu2024multistagelargelanguagemodel}, but balancing phonetic and semantic information under strict length constraints remains an open challenge.

\begin{figure}
    \centering
    \includegraphics[width=\linewidth]{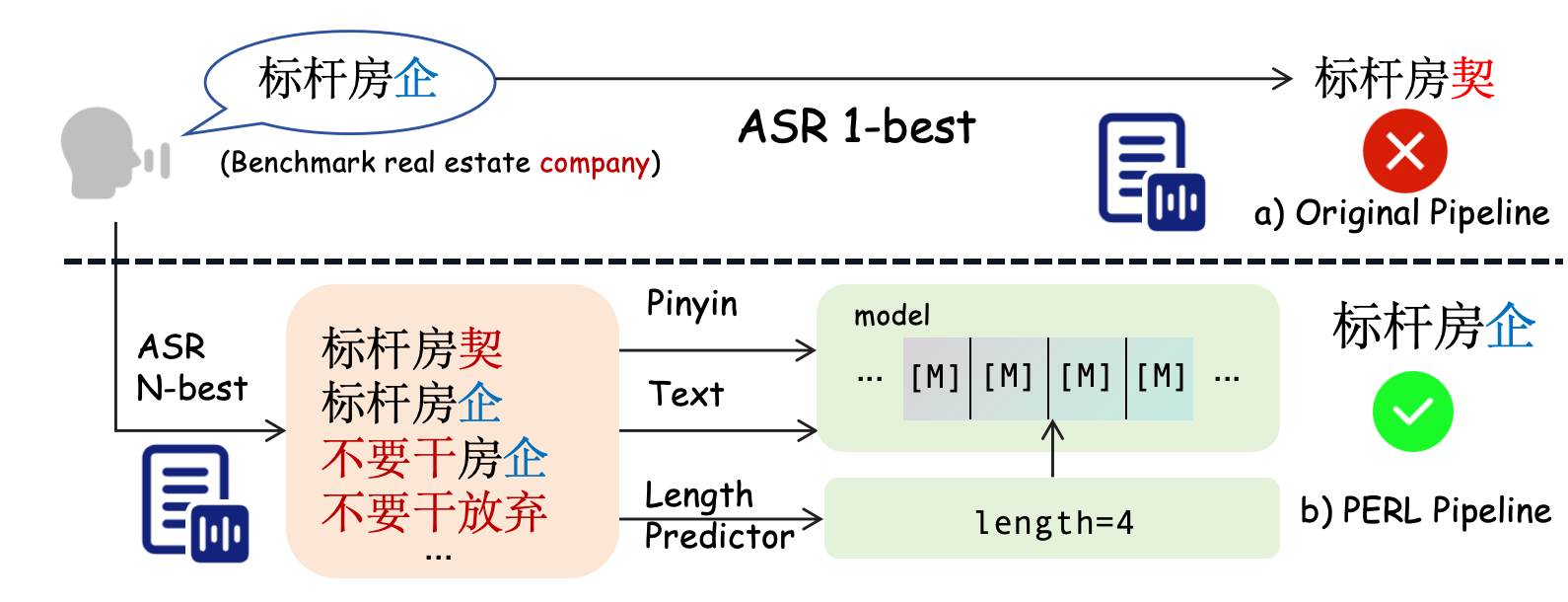}
    \caption{Comparison between the standard ASR pipeline (a) and our proposed PERL pipeline (b). Our proposed pipeline employs ASR N-best as the input and incoporates length information to determine masked tokens, and jointly employ Pinyin and Semantic information for error correction.}
    \label{fig:intro}
\end{figure}

To tackle these challenges, we propose the \textbf{Pinyin Enhanced Rephrasing Language (PERL)} pipeline (Figure~\ref{fig:intro}), which improves ASR correction by jointly modeling semantic and phonetic information. In addition, to comprehensively evaluate error correction in domain-specific scenarios, we introduce the \textbf{Do}main N-best \textbf{A}SR \textbf{D}ataset (DoAD). Our key contributions can be summarized as follows:

\begin{enumerate}
    \item We propose \textbf{PERL}, a constrained rephrasing framework for Chinese ASR N-best error correction that explicitly models the target length via a dedicated predictor and enforces it through mask-based generation.
    
    \item We design a \textbf{phonetic--semantic fusion mechanism} that integrates Pinyin-based phonological representations with contextual semantic embeddings via sentence-conditioned token-wise gating.
    
    \item We construct \textbf{DoAD}, a domain-specific N-best ASR benchmark with a high proportion of unequal-length cases, providing a challenging testbed for length-constrained correction.
\end{enumerate}

\begin{figure*}[t]
    \centering
    \includegraphics[width=0.8\linewidth]{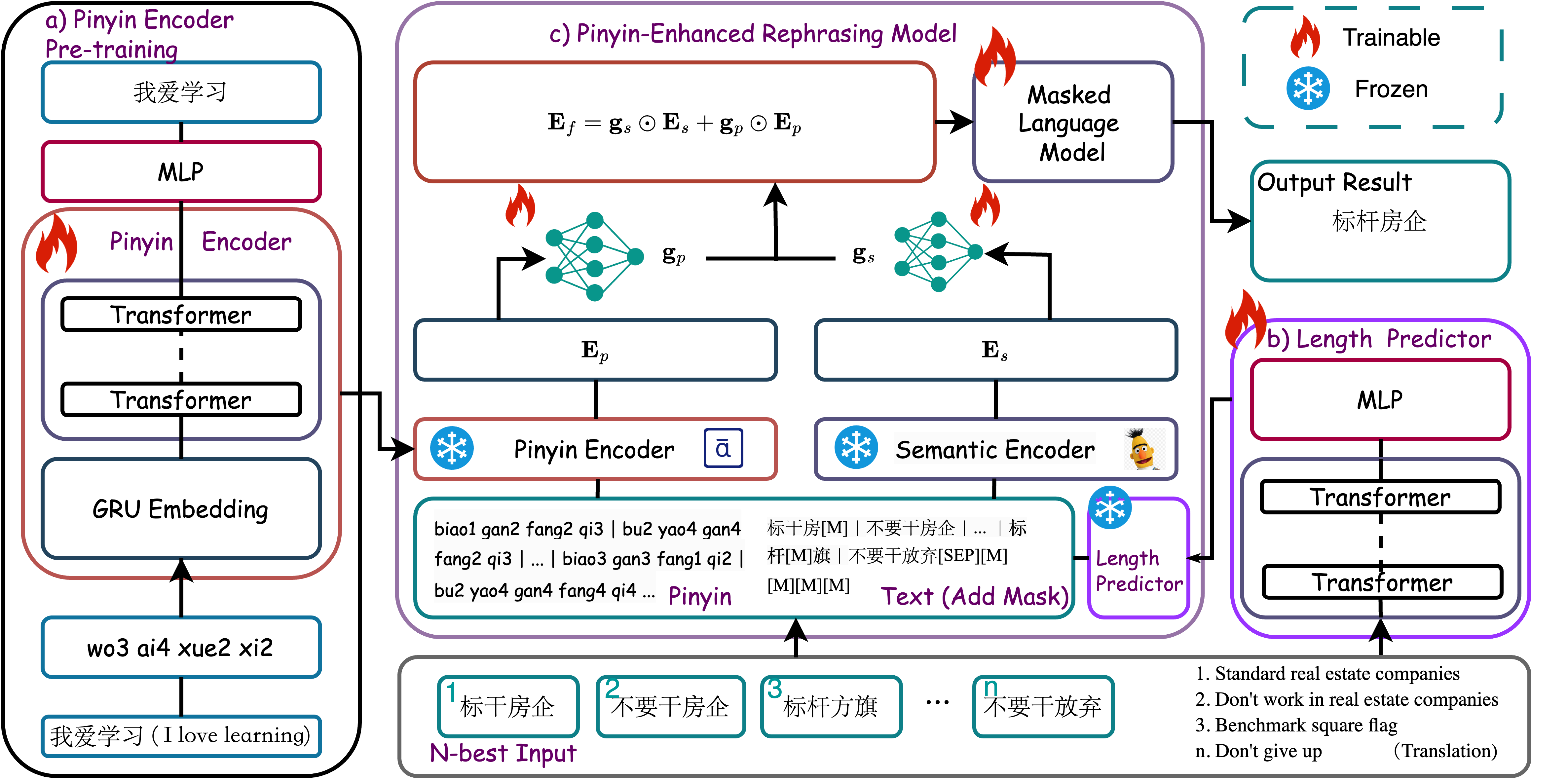}
    \caption{The overall architecture of the proposed model comprises three main components: (a) a pre-trained Pinyin encoder (left), which generates phonetic embeddings; (b) a length predictor (right), which estimates the target output length from the N-best inputs; and (c) a Pinyin-Enhanced Rephrasing Model (center), which integrates phonetic and semantic embeddings to perform masked token prediction. The Pinyin encoder and the length predictor are trained in the first two stages, after which their parameters are frozen and incorporated into the training of the rephrasing model.
}
    \label{fig:model}
\end{figure*}

\section{Method}
Our method consists of three main stages: 1) pretrain a Pinyin encoder to capture phonological information (Section \ref{sec:Pinyin encoder}), 2) pretrain a length predictor to handle the variable-length issue in ASR correction (Section \ref{sec:length_predict}), and 3) train a rephrasing model that integrates semantic and phonetic features for the final correction (Section \ref{sec:pinyin-enhanced_rephrasing_model}). Figure~\ref{fig:model} provides an overview.

\subsection{Pinyin Encoder}\label{sec:Pinyin encoder}
The Chinese Pinyin system provides a Latin-based transcription of Chinese characters that carries rich phonological information. To exploit this, we design a Pinyin encoder that maps a character sequence to phonetic embeddings. Given a sentence $x=(x_1,x_2,\dots,x_n)$ with $n$ characters, the encoder converts each character $x_i$ into a Pinyin sequence $p_{i,1},\dots,p_{i,l_i}$, where $l_i$ is the length of the transcription. The encoder then produces the phonetic representation $\mathbf{E}_p=\text{Enc}_p(x)$. It operates in three steps:

\paragraph{Pinyin embedding} Each token $p_{i,j}$ is assigned to a vector representation $\mathbf{p}_{i,j}$.

\paragraph{Sequential encoding} A uni-directional GRU encodes the Pinyin sequence
    $
        \hat{\mathbf{p}}_{i,j} = \text{GRU}(\hat{\mathbf{p}}_{i,j-1}, \mathbf{p}_{i,j}).
    $
    The last hidden state $\hat{\mathbf{p}}_{i,l_i}$ is used as the representation of the character $x_i$, forming the sequence
    $
    \mathbf{E}_{p,0} = (\hat{\mathbf{p}}_{1,l_1}, \hat{\mathbf{p}}_{2,l_2},\dots,\hat{\mathbf{p}}_{n,l_n})
    $.
    
\paragraph{Global encoding} To capture long-range dependencies, we apply $L_p$ Transformer layers 
    $
    \mathbf{E}_{p,l} = \text{Transformer}( \mathbf{E}_{p,l-1}), \quad l \in [1,L_p]
    $
    
The final phonetic representation $\mathbf{E}_{p,L_p}$ is projected to predict the corresponding Chinese characters using learnable parameters $\mathbf{W}^p$ and $\mathbf{b}^p$. For position $i$, the prediction is defined as
\[
\hat{y}_i = \text{Softmax}(\mathbf{W}^p \mathbf{E}_{p,L_p,i} + \mathbf{b}^p), \quad \mathbf{E}_{p,L_p,i} \in \mathbf{E}_{p,L_p}.
\]

The encoder is pre-trained with a cross-entropy loss, where $y_i$ denotes the ground-truth character at position $i$:
$$
\mathcal{L}_p = -\tfrac{1}{n} \sum_{i=1}^{n} y_i \log \hat{y}_i.
$$

\subsection{Length Predictor}\label{sec:length_predict}
Chinese N-best ASR correction is a \emph{length-constrained} rephrasing problem: the corrected output may be shorter or longer than any single hypothesis, and an incorrect length budget can propagate errors to the downstream mask-filling stage. We therefore introduce a dedicated length predictor to estimate the target character length from the N-best list.

\paragraph{Input}
Given an N-best list $\{S_k\}_{k=1}^{n}$, we concatenate hypotheses with a delimiter ``$|$'':
\[
S_{\text{concat}} = S_1 \,|\, S_2 \,|\, \dots \,|\, S_n .
\]
The ground-truth label is the character length $l$ of the reference sentence. We set the maximum length to $L$ (e.g., $L{=}128$) and clamp labels into $[1,L]$.

\paragraph{Ordinal length modeling}
A flat multi-class classifier over lengths can be brittle when the test-time length distribution shifts across domains. To improve generalization, we model length as an \emph{ordinal} variable. Specifically, we encode $S_{\text{concat}}$ using a BERT encoder and take the [CLS] embedding $\mathbf{h}\in\mathbb{R}^{d}$:
\[
\mathbf{h} = \text{Enc}_s(S_{\text{concat}}).
\]
For each threshold $m\in\{1,\dots,L\}$, the predictor estimates the cumulative probability
\[
p_m = P(l \geq m \mid S_{\text{concat}}) = \sigma(\mathbf{w}_m^\top \mathbf{h} + b_m),
\]
where $\sigma(\cdot)$ is the sigmoid function. The supervision signal for the threshold $m$ is
$
y_m=\mathbbm{1}[l \geq m].
$
We train the predictor using a sum of binary cross-entropy losses across thresholds:
\[
\mathcal{L}_{\text{len}} = - \sum_{m=1}^{L}\Big( y_m \log p_m + (1-y_m)\log(1-p_m)\Big).
\]
This formulation explicitly exploits the natural ordering of length values and is less sensitive to unseen lengths than a flat softmax classifier.

\paragraph{Inference with length uncertainty}
Instead of relying on a single point estimate, we decode with a small candidate set of lengths to mitigate prediction uncertainty.
We first recover a discrete distribution over lengths by
\[
P(l=m) = 
\begin{cases}
1-p_{2}, & m=1,\\
p_{m}-p_{m+1}, & 1<m<L,\\
p_{L}, & m=L,
\end{cases}
\]
and take the top-$K$ lengths $\mathcal{C}=\{l^{(1)},\dots,l^{(K)}\}$ 
For each candidate length $l^{(k)}$, we run the downstream rephrasing model (Section~\ref{sec:pinyin-enhanced_rephrasing_model}) with $l^{(k)}$ appended masks, and select the final output with the lowest loss in masked language modeling (i.e., the highest likelihood in masked positions). This adds only a small overhead, while improving robustness under the length distribution shift.

\subsection{Pinyin-Enhanced Rephrasing Model} \label{sec:pinyin-enhanced_rephrasing_model}
The final correction model integrates semantic and phonetic information while incorporating the predicted length. It consists of two stages: input processing and feature encoding/prediction.

\paragraph{Input processing} The N-best hypotheses are concatenated into $S_{\text{concat}}$. The frozen length predictor is then used to estimate the target length $\hat{l}$ (Section~\ref{sec:length_predict}). To align with this prediction, $\hat{l}$ mask tokens are appended to $S_{\text{concat}}$. Following \cite{liu2024chinesespellingcorrectionrephrasing}, we further apply a dynamic masking strategy, where a subset of tokens in $S_{\text{concat}}$ is randomly replaced with the [M] symbol, yielding the masked sequence $S_{\text{masked}}$.

\paragraph{Feature encoding} The masked sequence $S_{\text{masked}}$ is processed by the semantic encoder, while the concatenated sentence $S_{\text{concat}}$ is passed through the Pinyin encoder; the parameters for the Pinyin encoder in Section \ref{sec:Pinyin encoder} will be frozen.

\[
\mathbf{E}_s = \text{Enc}_s(S_{\text{masked}}), \quad
\mathbf{E}_p = \text{Enc}_p(S_{\text{concat}}).
\]
The sentence-level semantic embedding is obtained by averaging token embeddings over the maximum sequence length:
\[
    \bar{\mathbf{e}}_s = \tfrac{1}{l_{\max}} \sum_{i=1}^{l_{\max}} \mathbf{E}_s[i],
\]
where $l_{\max}$ denotes the maximum sentence length and $\mathbf{E}_s[i]$ is the semantic embedding of the $i$-th token.

\paragraph{Feature fusion} 
For each token, semantic and phonetic features are combined through learned gating weights:
\[
\begin{aligned}
g_{s,i} &= \sigma\!\left(\text{MLP}_s\big([\mathbf{E}_s[i]; \mathbf{E}_p[i]; \bar{\mathbf{e}}_s]\big)\right), \\
g_{p,i} &= \sigma\!\left(\text{MLP}_p\big([\mathbf{E}_s[i]; \mathbf{E}_p[i]; \bar{\mathbf{e}}_s]\big)\right).
\end{aligned}
\]
where $[\cdot;\cdot]$ denotes vector concatenation, $\text{MLP}_s$ and $\text{MLP}_p$ are two-layer perceptrons with ReLU activation, and $\sigma(\cdot)$ is the sigmoid function.

The final fused embedding is computed as
\[
\mathbf{E}_f[i] = g_{s,i}\mathbf{E}_s[i] + g_{p,i}\mathbf{E}_p[i].
\]

\paragraph{Prediction} The fused representation $\mathbf{E}_f$ is passed to a mask prediction model to predict masked tokens. The model is trained with a cross-entropy objective over masked positions:
\[
\mathcal{L}_{\text{mask}} = -\sum_{i \in \mathcal{M}} y_i \log \hat{y}_i,
\]
where $\mathcal{M}$ denotes the set of masked indices and $y_i$ is the ground-truth label.


\section{Experimental Setup}

\subsection{Data}\label{sec:data}

We perform experiments on the ChineseHP/Aishell-1 (CHP/Aishell-1) dataset \cite{tang_pinyin_2024} and construct \textbf{the} \textbf{Domain} \textbf{ASR} dataset (DoAD). The construction process is summarized as follows:

\paragraph{Data Processing} We start with ECSpell data \cite{lv_general_2023}, a text domain error correction dataset. Each gold-standard sentence is segmented by punctuation and normalized to remove Arabic numerals and punctuation marks.  

\paragraph{Text-to-Speech} Normalized sentences are converted into speech using Microsoft Azure\footnote{\url{https://learn.microsoft.com/zh-cn/azure/ai-services/speech-service/text-to-speech}}. To simulate noisy conditions, white noise between 10~dB and 20~dB is added randomly.  

\paragraph{ASR Generation} A Whisper ASR model \cite{radford2022robustspeechrecognitionlargescale} is used to generate N-best hypotheses. Specifically, we adopt the Belle-distilwhisper-large-v2-zh model\cite{BELLE}.  

Table~\ref{tab:asr_correction_data} reports dataset statistics. Compared to Aishell-1, DoAD contains a much larger proportion of unequal-length sentences, which poses additional challenges for LLMs to recover the correct sequence length. We attribute this to the noise, which makes ASR correction more difficult than in clean speech. Although DoAD is synthetically constructed using TTS and noise injection, it enables controlled generation of unequal-length ASR errors. However, it may not fully capture real-world phenomena such as disfluencies, accents, or spontaneous speech patterns, which we leave for future work.

\begin{table}[ht!]
\centering
\caption{Dataset statistics. \#Sen denotes the number of sentences, \#Len the average sentence length, and \#Equal the number of sentences where the 1-best output matches the reference length. Both training and test sets are reported.}
\renewcommand{\arraystretch}{0.9}

\begin{tabular}{lrrr}
\toprule
Dataset & \#Sen (Train/Test) & \#Len & \#Equal \\
\midrule
DoAD-Law   & 4,040 / 1,010  & 13.69 / 13.64  & 2,469 / 263 \\
DoAD-Med   & 10,978 / 1,820 & 11.30 / 11.55  & 5,878 / 1,032 \\
DoAD-Odw   & 5,137 / 1,480  & 12.45 / 12.41  & 2,502 / 696 \\
CHP/Aishell1 & 120,099 / 7,176 & 14.41 / 14.60 & 113,039 / 6,699 \\
\bottomrule
\end{tabular}
\label{tab:asr_correction_data}
\end{table}

\subsection{Experiment Setup and Evaluation}
The experiments are conducted on two NVIDIA 3090 GPUs. The Pinyin encoder is pre-trained on the Wang271k \cite{wang-etal-2018-hybrid} gold-labeled dataset, with 10,000 sentences held out for validation. In principle, any error-free Chinese corpus could be used for this pre-training. The encoder is trained for 10 epochs with an initial learning rate of $5 \times 10^{-4}$. The Aishell-1 and DoAD datasets are used to train the length predictor, starting with a learning rate $2\times 10^{-5}$.
Finally, PERL training follows the same training parameters shown in \cite{liu2024chinesespellingcorrectionrephrasing}. The mask rate is set to 0.2, and the maximum length is 128. We choose the BERT model \cite{devlin2018bert} as the semantic encoder and select the 5-best results as input for each scenario. 

For evaluation, we adopt Character Error Rate (CER) and Character Error Rate Reduction (CERR). CER is computed as the edit distance between the predicted and reference sequences normalized by the reference length, while CERR measures the relative reduction of CER compared to the best ASR baseline.

\subsection{Baselines}
To thoroughly investigate the effectiveness of our proposed model, we select several representative pretrained language models (PLMs) and large language models (LLMs) as baselines. For PLMs, we employ \textbf{MacBERT} \cite{cui-etal-2020-revisiting}, a Chinese pre-trained model that improves on BERT by using a masked correction pretraining strategy, where original tokens are replaced with semantically similar words instead of the special mask symbol \cite{pycorrector}. We also include \textbf{T5} \cite{raffel2023exploringlimitstransferlearning}, a unified text-to-text Transformer that reformulates every NLP problem as a sequence-to-sequence task, allowing flexible adaptation to classification, summarization, and error correction.  In addition, we use \textbf{ReLM} \cite{liu2024chinesespellingcorrectionrephrasing}, the Rephrasing Language Model trained with a mask filling objective, to rephrase entire sentences rather than detecting errors token by token, emphasizing fluency and natural rewording. 

For LLMs, we benchmark against \textbf{GPT-4o}, a generative model developed by OpenAI and optimized from the GPT-4 architecture with state-of-the-art reasoning ability, \textbf{DeepSeek-V2.5}, which demonstrates strong capabilities in coding and mathematical reasoning, and \textbf{Qwen 2.5}, a large multilingual model pre-trained on diverse corpora. For generative baselines, we use a zero-shot prompt setup following \cite{tang_pinyin_2024}, where the N-best hypotheses are provided as input and the model is instructed to generate a corrected sentence. 
We do not apply additional constrained decoding or length control for LLMs.

\section{Results}
\subsection{N-best ASR Correction}
\begin{table}[t!]
\centering

\caption{Performance of our post-correction model under different ASR systems. $O_{cp}$ and $O_{nb}$ denote the character-level oracle and the N-best selection oracle, respectively. The best results are shown in \textbf{bold}.}
\scalebox{0.86}{
\begin{tabular}{lllll}
\toprule
\multirow{2}{*}{Method} & \multicolumn{4}{c}{\textbf{CER\%}$\downarrow$ $_{-\text{CERR}\%\downarrow}$} \\ \cline{2-5} 
                        & CHP/Aishell1 & DoAD-Law & DoAD-Med & DoAD-Odw \\ \hline
Baseline & 5.84 & 12.64 & 16.09 & 15.94 \\ \hline
$O_{cp}$ & 2.93{\color{teal}$_{-49.83}$} & 3.83{\color{teal}$_{-69.70}$} & 6.13{\color{teal}$_{-61.90}$} & 4.41{\color{teal}$_{-73.59}$} \\ 
$O_{nb}$ & 3.49{\color{teal}$_{-32.70}$} & 9.44{\color{teal}$_{-25.31}$} & 12.66{\color{teal}$_{-28.37}$} & 11.89{\color{teal}$_{-28.80}$} \\ 
\midrule
MacBERT & 5.06{\color{blue}$_{-13.36}$} & 13.16{\color{blue}$_{+4.11}$} & 17.28{\color{blue}$_{+7.40}$} & 16.76{\color{blue}$_{+5.14}$} \\
T5      & 5.49{\color{blue}$_{-6.00}$}  & 14.34{\color{blue}$_{+13.45}$} & 18.52{\color{blue}$_{+15.10}$} & 18.49{\color{blue}$_{+16.00}$} \\
ReLM    & 5.20{\color{blue}$_{-10.96}$} & 5.21{\color{blue}$_{-58.78}$} & 6.91{\color{blue}$_{-57.05}$} & 6.17{\color{blue}$_{-61.23}$} \\ 
\midrule
GPT-4o  & 4.41{\color{blue}$_{-24.49}$} & 9.83{\color{blue}$_{-22.23}$} & 13.27{\color{blue}$_{-17.53}$} & 12.40{\color{blue}$_{-22.20}$} \\ 
DeepSeek & 4.25{\color{blue}$_{-27.22}$} & 9.12{\color{blue}$_{-27.85}$} & 12.61{\color{blue}$_{-21.63}$} & 11.68{\color{blue}$_{-26.72}$}\\   
Qwen2.5 & 4.22{\color{blue}$_{-27.74}$} & 8.28{\color{blue}$_{-34.49}$} & 12.16{\color{blue}$_{-24.43}$} & 11.15{\color{blue}$_{-30.05}$} \\
\midrule
PERL    & \textbf{4.10{\color{blue}$_{-29.11}$}} & \textbf{3.41{\color{blue}$_{-73.02}$}} & \textbf{3.91{\color{blue}$_{-75.70}$}} & \textbf{4.87{\color{blue}$_{-69.44}$}} \\
\bottomrule
\end{tabular}}

\label{tab:cer_results}
\end{table}

We utilize the ChineseHP \cite{tang_pinyin_2024} dataset for Aishell-1 and perform our own ASR on distil-whisper\cite{BELLE}. Table \ref{tab:cer_results} shows the N-best ASR correction result, where CER represents the average character error rate, and CERR represents the percentage decrease in the average character error rate. $o_{cp}$ represents the best character combination oracle and $o_{nb}$ represents the best selection oracle. 

Under our experimental framework, the baseline model exhibited relatively high CER in the law, medical, and official document domains of the DoAD dataset compared to Aishell-1, highlighting the challenge posed to our model. Empirical results show that PERL substantially reduces CER across all datasets, achieving a 29.1\% reduction on Aishell-1 and around 70\% across different domains on DoAD. Notably, the improvement on DoAD can outperform a character-level oracle that is restricted to selecting characters appearing in the N-best list, suggesting PERL benefits from contextual rephrasing beyond per-position selection. This gain can be attributed to the ability of PERL to explore a broader semantic space and take advantage of contextual dependencies, which is particularly effective when the DoAD beam search produces low-quality hypotheses, demonstrating the robustness of our proposed framework.

In contrast, LLMs perform worse than PERL in DoAD. We attribute this to the inherent limitations of generative models when handling tasks with length constraints, where noisy inputs often mislead them into producing erroneous output.


\subsection{Result Analysis}
\begin{figure}
    \centering
    \includegraphics[width=\linewidth]{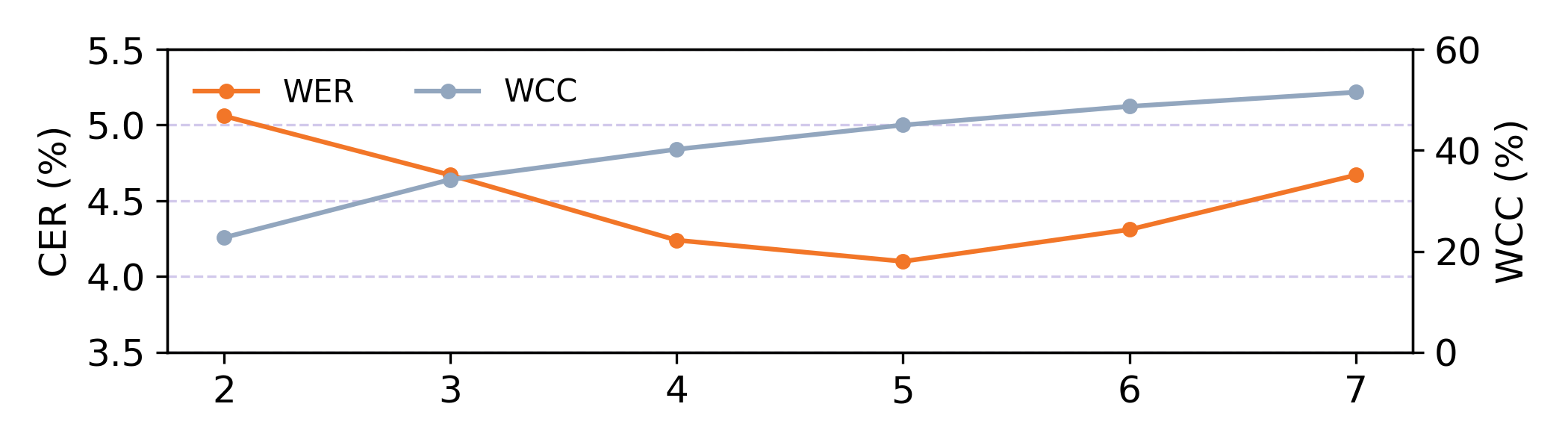}
    \vspace{-2em}
    \caption{Character Error Rate (CER) and Wrong Character Coverage (WCC) across different $n$-best settings. The left y-axis shows CER, while the right y-axis shows WCC.}
    \label{fig:wcc_wer}
    
\end{figure}

\textbf{The choice of $n$.}  
For each erroneous character in the 1-best hypothesis, we calculate the proportion of $n$-best candidates that contain its correct counterpart. We define this metric as \emph{Wrong Character Coverage (WCC)}, which measures how often the correct character appears within the $n$-best list. Figure~\ref{fig:wcc_wer} reports the WCC performance of our proposed PERL model under different $n$ settings on the ChineseHP/Aishell-1 dataset. We observe that the Character Error Rate (CER) reaches its minimum when $n=5$. Interestingly, although WCC continues to increase to 6, the CER worsens. This suggests that simply expanding $n$ does not guaranty a better correction and that effectiveness depends on the ability of the model to leverage additional hypotheses and processing long inputs.

\textbf{ASR Model Selection.} To assess the robustness of our model across different ASR systems, we evaluate it on Belle-distilwhisper-large-v2-zh, as detailed in Section \ref{sec:data}, as well as on whisper-small and whisper-large-v3 to process the DoAD speeches. Specifically, we measure the CER of the PERL on these systems. The results in Table \ref{tab:whisper_models} demonstrate that PERL substantially reduces CER compared to the 1-best baselines, highlighting its strong robustness in various ASR models.

\begin{table}[t!]
    \centering

    \caption{CER of whisper models across different domains of DoAD.}
    \label{tab:whisper_models}

    \begin{tabular}{llcccc}
        \toprule
        Model & Domain & Baseline & PERL & $o_{nb}$ & $o_{cp}$ \\
        \midrule
        \multirow{3}{*}{whisper-large-v3} & law & 15.81 & 5.65 & 9.24 & 7.67 \\
                                          & med & 15.86 & 7.20 & 10.37 & 9.15 \\
                                          & odw & 14.75 & 7.61 & 9.09 & 8.04 \\
        \midrule
        \multirow{3}{*}{whisper-small}    & law & 26.76 & 10.77 & 18.01 & 14.72 \\
                                          & med & 29.69 & 13.22 & 20.82 & 17.81 \\
                                          & odw & 30.02 & 12.82 & 19.93 & 16.51 \\
        \bottomrule
    \end{tabular}

\end{table}

\textbf{Length Prediction.} To assess the significance of the length prediction module, we evaluate our model's length correction performance, as shown in Table \ref{tab:length_predict_result}. The pre-trained length predictor demonstrates improved accuracy in predicting the correct length compared to relying solely on the 1-best result. This enhancement enables the model to determine the masked tokens, improving overall prediction.

\begin{table}[t!]
\centering
\caption{Length correction on different Whisper models.}
\label{tab:length_predict_result}

\begin{tabular}{llccc}
\toprule
\multirow{2}{*}{Model} & \multirow{2}{*}{Method} & \multicolumn{3}{c}{\#Equal} \\
\cmidrule(lr){3-5}
 & & Law & Med & Odw \\
\midrule
\multirow{2}{*}{whisper-large-v3} 
  & 1-best          & 910  & 1641 & 1393 \\
  & length predict  & 954  & 1685 & 1410 \\
\midrule
\multirow{2}{*}{whisper-small}    
  & 1-best          & 853  & 1551 & 1289 \\
  & length predict  & 913  & 1650 & 1310 \\
\bottomrule
\end{tabular}
\end{table}

\textbf{Ablation Experiments.}
We evaluate PERL on the Aishell-1 and DoAD datasets under different ablation settings as shown in Table \ref{tab:abla}: (1) removing the length predictor and using only the length of the 1-best prediction, (2) removing the Pinyin encoder, and (3) using only the 1-best prediction during correction. Our findings show that removing the Pinyin encoder or relying solely on the N-best results significantly degrades performance. Furthermore, removing the length predictor during correction results in even more significant performance degradation, as the language model needs to rephrase the sentence into shorter or longer sequences.

\textbf{Case Study.} ASR N-best Correction example is shown in Table \ref{tab:case}, PERL could calculate the correct length and incorporate Pinyin knowledge to recover the wrong token \chinese{\textcolor{red}{松}} (\pinyin{sōng}, pine) into \chinese{\textcolor{blue}{宗}} (\pinyin{zōng}, sect) as they share similar Pinyin. At the same time, the ReLM model failed by correcting this in \chinese{\textcolor{orange}{块}} (\pinyin{kuài}, piece) since it cannot use the Pinyin information, and the 1-best result is unable to provide potential corrections compared to the N-best results.
\begin{table}[t!]
\vspace{-1em}
\centering
\caption{Example of ASR N-best correction for different models, where we use \textcolor{red}{red}/\textcolor{blue}{blue}/\textcolor{orange}{orange} colors to represent \textcolor{red}{wrong}/\textcolor{blue}{correct}/\textcolor{orange}{wrong edit} tokens, best viewed in color.}

\begin{tabular}{|l|l|}
\hline
\textbf{5-Best Sentences} & \textbf{Model Outputs} \\ \hline
\begin{tabular}[c]{@{}l@{}} 
\chinese{目前挂牌的只有几\textcolor{red}{松}土地} \\ 
\chinese{目前挂\textcolor{red}{排}的只有几\textcolor{red}{松}土地} \\ 
\chinese{目前挂牌的只有几\textcolor{red}{松}土} \\ 
\chinese{目前挂牌的只有几\textcolor{red}{宋}土地} \\ 
\chinese{目前挂牌的只有几\textcolor{blue}{宗}土地} 
\end{tabular} 
& 
\begin{tabular}[c]{@{}l@{}} 
\chinese{目前挂牌的只有几\textcolor{blue}{宗}土地} (PERL) \\ 
\chinese{目前挂牌的只有几\textcolor{red}{松}土地} (1-best)\\ 
\chinese{目前挂牌的只有几\textcolor{orange}{块}土地} (ReLM)\\ 
\chinese{目前挂牌的只有几\textcolor{blue}{宗}土地} (GPT-4o)\\ 
\chinese{目前挂牌的只有几\textcolor{blue}{宗}土地} (Gold)\\ 
\end{tabular} \\ \hline
\multicolumn{2}{|l|}{Translation: Currently, only a few sects of land are listed for sale.} \\ \hline
\end{tabular}

\label{tab:case}
\end{table}

\textbf{Latency Analysis.} Beyond correction accuracy, inference efficiency is also evaluated, a key requirement for deployment in real-world ASR scenarios. On a single NVIDIA 3090 GPU, lightweight encoder-based PLMs achieve the fastest speeds per input (MacBERT: 1.67 ms, ReLM: 2.87 ms). The PERL pipeline introduces only minor overhead (3.09 ms) while lowering CER, offering a strong balance between speed and accuracy. In contrast, large generative LLMs are substantially slower (hundreds of ms) and thus less practical for latency-sensitive ASR correction.

\begin{table}[t!]
\centering

\caption{Ablation of PERL model on ChineseHP/Aishell-1 and DoAD datasets with different settings.}

\begin{tabular}{r|llll}
\hline
\toprule
\textbf{Method} & \textbf{CHP/Aishell-1} & \textbf{Law} & \textbf{Med} & \textbf{Odw} \\ \midrule
PERL            & 4.10                        & 3.41              & 3.91               & 4.87               \\ 
- w/o len         & 7.28                         &     5.08                &         5.47            &       7.12              \\ 
- w/o pho         & 4.78                        & 4.27                &         4.67       & 5.83       \\ 
- w/o N-best     & 4.31                      & 3.69          & 4.62                & 5.15                 \\ 
\bottomrule
\end{tabular}

\label{tab:abla}

\end{table}


\section{Conclusion}
PERL integrates a Pinyin module to reduce phonetic errors and a length predictor to address sequence inconsistencies in N-best Chinese ASR error correction. To evaluate performance under challenging conditions, we introduce the highly noisy DoAD dataset. Experiments demonstrate that PERL effectively leverages semantic and phonetic features to correct ASR errors while maintaining low latency, making it both accurate and practical for real-world applications.

\bibliographystyle{IEEEtran}
\bibliography{refs}

\end{document}